# Can we Improve Prediction of Psychotherapy Outcomes Through Pretraining With Simulated Data?

Niklas Jacobs[1, 2], Manuel C. Voelkle[3], Norbert Kathmann[2], and Kevin Hilbert[1]

**Abstract**

In the context of personalized medicine, machine learning algorithms are growing in popularity. These algorithms require substantial information, which can be acquired effectively through the usage of previously gathered data. Open data and the utilization of synthetization techniques have been proposed to address this. In this paper, we propose and evaluate an alternative approach that uses additional simulated data based on summary statistics published in the literature. The simulated data are used to pretrain random forests, which are afterwards fine-tuned on a real dataset. We compare the predictive performance of the new approach to random forests trained only on the real data. A Monte Carlo Cross Validation (MCCV) framework with 100 iterations was employed to investigate significance and stability of the results. Since a first study yielded inconclusive results, a second study with improved methodology (i.e., systematic information extraction and different prediction outcome) was conducted. In Study 1, some pretrained random forests descriptively outperformed the standard random forest. However, this improvement was not significant ($t(99) = 0.89$, $p = 0.19$). Contrary to expectations, in Study 2 the random forest trained only with the real data outperformed the pretrained random forests. We conclude with a discussion of challenges, such as the scarcity of informative publications, and recommendations for future research.

## Introduction

Within the advances towards personalized treatment for patients, prediction of treatment outcomes poses an important challenge to, for example, guide the assignment of the most promising therapy approach (e.g., DeRubeis et al., 2014). Machine learning (ML) algorithms play an important role in this context, because they provide high predictive performance and are capable of handling high dimensional, complex data (Meinke et al., 2024). For ideal performance of ML, large datasets are required. Unfortunately, in many fields, especially clinical psychology and medicine, large open access datasets are rarely available due to ethical and data protection reasons. As an alternative to sharing real data, synthetic datasets have been proposed (for an introduction see Goncalves et al., 2020). Such synthetic datasets capture the underlying structure (e.g., linear/nonlinear associations, clustering) of the real data but reduce the risk for identification of participants. Inconveniently, identifying participants is not impossible but only difficult (Hittmeir et al., 2020), which might be one reason this practice is not yet commonly applied in clinical psychology.

---

[1] Department of Psychology, Health and Medical University Erfurt, Germany

[2] Department of Psychology, Humboldt-Universität zu Berlin, Germany

[3] Department of Psychology, University de Fribourg, Germany

As an alternative to the aforementioned approaches, we investigated whether published descriptive statistics from previous studies, derived from diverse populations and contexts cross-nationally, can be utilized to improve predictive performance. Ideally this could reduce both overfitting and sampling bias for the pretrained algorithm. This is the case because noise in the data, which leads to overfitting, could be averaged out through the integration of heterogenous sample characteristics for simulation. Sampling bias on the other hand, is the result of characteristics only apparent in certain subgroups. Since the studies used for simulation stem from different countries, populations and time points, we expected sample specific characteristics to be of decreased distinctiveness in the simulated data. Pretraining is common, especially in the application of deep neural networks where it is known to improve robustness and reduce overfitting (Hendrycks et al. 2019).

## Study 1: Improving the prediction of response to psychotherapy

In this study, we implemented the proposed approach by gathering descriptive information for selected variables between treatment outcome groups (e.g., responders/non-responders), and used those to simulate new cases for algorithms to be trained on. After simulation, we compared ML algorithms trained on both simulated and real data (*pretrained* algorithm) with a ML algorithm trained only on the real data (*standard* algorithm). Additionally, the weight given to the simulated data was varied and differences analyzed. To test this approach, we predicted whether patients with obsessive compulsive disorder (OCD) would respond to cognitive behaviour therapy.

We hypothesized that the best performing pretrained algorithm would display significantly higher performance as compared to the standard algorithm. Additionally, we hypothesized that the pretrained algorithm with low weight for the simulated data would perform superior as compared to the other pretrained approaches. This was done because we expected that the simulated information would be especially helpful when utilized in "uncertain" cases (i.e., when the fine-tuned model, trained on the real data, has low confidence in its predictions).

## Methods

### Sample

The dataset was previously utilized for ML based outcome prediction by Hilbert et al. (2021). It consisted of 463 patients (56% female) from the outpatient clinic of the Humboldt-Universität zu Berlin. Inclusion criteria were an obsessive compulsive disorder (OCD) diagnosis, determined through the Structured Clinical Interview for DSM-IV Axis I disorders (First et al., 1997), age between 18 and 70 years and a score >16 on the Yale-Brown obsessive-compulsive scale (YBOCS, Goodman, 1989). Patients were excluded if they showed prominent suicidal ideation, any lifetime substance dependence, a borderline personality disorder, or a comorbid psychotic disorder. All patients received CBT. Analysis of routinely collected data met the ethical standards of the revised Declaration of Helsinki. All participants provided written informed consent that the data being collected during their therapies will be used for research and might be published.

**Simulation of data**

While the dataset contained 504 variables, we aimed to simulate only a subset of six interesting variables (YBOCS; BDI-II (Beck et al., 1996), GAF (APA, 1994), OCI-R (Foa et al., 2022), age and age at onset of disorder) which we expected to be important for prediction based on the analyses by Hilbert et al. (2021). We sampled the simulated data from two multivariate normal distributions (MVND) $N(\mu, \Sigma)$ where $\mu$ is a mean vector and $\Sigma$ a variance-covariance matrix. One MVND was defined for responder patients, the other for non-responder patients. Therefore, mean values, standard deviations and variable intercorrelations for responders and non-responders were searched for via google scholar and PubMed (for details see appendix 1).

Since simulating only six variables considerably limited dimensionality and complexity of the simulated dataset, as compared to the real data, we employed a second, extended, simulation procedure. This procedure simulated all other features in the real dataset in addition to our six variables of interest. For these additional features we did not obtain mean and (co-)variance estimates from the literature, but simulated the additional features based on their descriptive statistics in the train split. This allowed to maintain the original structure of the real data while also including the literature based information for the six variables of interest.

As an exploratory analysis, the MADRS (Montgomery & Åsberg, 1979) was added to the features of interest simulation to examine the effect of additional simulated features. For the MADRS, no study met all criteria for inclusion (see appendix 1), which is why we opted for this exploratory addition to the simulation.

In the end, we simulated 4 datasets.

1. *Six-features*: In this dataset we simulated only the six variables of interest.
2. *Seven-features*: In this dataset we simulated the six features of interest and the MADRS as an exploratory addition.
3. *All-features*: In this dataset all features were simulated, the six features of interest based on the literature, and all others based on the train set of the data.
4. *All-features MADRS included*: In this dataset all features were simulated, the six features of interest and the MADRS on the basis of the literature, and all other features based on the train set of the data.

In the following analyses, these datasets were utilized separately for pretraining the random forests.

**Pretraining**

The simulated data was used to create a random forest by training a portion of its ensembled decision trees on the simulated data (pretraining) while the other portion was trained on the real dataset (fine-tuning). When provided with unseen real data, pretrained trees therefore used patterns derived from the literature to predict response, while fine-tuned trees utilized the associations in the real data. The single probabilistic predictions from each tree were then averaged into a final prediction. This final prediction was used to calculate the performance measures for the pretrained approaches. Furthermore, the weight given to the simulated data was also manipulated by varying the number of trees in pretraining as compared to the fine-tuning process.

Specifically, 20%, 50% and 100% as many pretrained as fine-tuned trees were examined (i.e., if 200 trees were trained in total, the 20% condition pretrained 33 and fine-tuned 167 trees). This was done to explore whether the pretraining is only helpful if the fine-tuned classifier is uncertain in its prediction. While the 20% option can only influence the final prediction in cases with an uncertain fine-tuning, the 100% option can alter the final prediction even if the fine-tuning trees consistently predict one class with high probabilistic predictions. In case the simulated data was always an improvement, even if the fine-tuned trees were confident in their prediction, the 100% option would be expected to obtain the best results.

**Machine learning pipeline**

Before conducting any calculation, the real dataset was randomly split into a training and a test set. Data preprocessing consisted of one hot encoding for categorical features, imputation on the basis of the train split via multiple imputation by chained equations (Raghunathan et al., 2000) and creation of the label variable indicating whether a case responded. Response was operationalized via the reliable change index (Jacobson & Truax, 1991). Following that, hyperparameters (e.g., number of trees in the random forest, number of nodes in a decision tree) were tuned using a grid search (i.e., assessing algorithm performance for all selected hyperparameter combinations). This hyperparameter-tuning was conducted separately for all datasets (real and simulated) since each dataset could have potentially required different hyperparameters for optimal classifier performance. Since the real dataset contained more responders than non-responders, it was balanced via SMOTENC (Lemaître et al., 2017), which is an approach to add synthetic cases that resemble the minority class to the dataset until both labels are equally prevalent. After the data was preprocessed, the random forests were fitted with differing numbers of decision trees trained on the simulated data. For more information on the pipeline see Appendix 2.

**Evaluation**

Once the algorithms were trained, they were applied to the previously unseen test cases to evaluate their predictive performance. We analyzed differences in balanced accuracy between pretrained and non-pretrained random forests as our primary outcome and exploratively computed the area under the receiver operator characteristic curve, sensitivity and specificity. The process of data simulation, machine learning analysis and calculation of performance metrics was repeated 100 times with varying random seeds to evaluate whether improvements were due to chance.

The balanced accuracy of the best performing pretrained algorithm was compared with the best performing non pretrained algorithm using the corrected resampled t test (Bouckaert & Frank, 2004; Nadeau & Bengio, 2003). The analysis was preregistered via OSF (osf.io/abkrq) and the Python scripts can be found on Github (https://github.com/jacobnikl/Pretraining-a-random-forest---Materials)

<div align="center">Results</div>

The performance metric means over iterations showed some variation between the different approaches. For performance metrics of approaches trained or fine-tuned on the balanced dataset refer to Table 1 (for results on other datasets/approaches see Tables A3/A4). The random forest without pretraining reached an average balanced accuracy of 0.526 with a standard deviation of 0.035. When pretraining was applied, the predictive performance improved. The strongest improvement occurred for the random forest pretrained on the data that simulated all features and assigned a weight of 50% to the simulated data. Its average balanced accuracy was 0.549 with a standard deviation of 0.052. Despite the descriptive improvements for the pretrained forests, this difference between the balanced standard approach and the best-performing balanced pretrained approach of 2.2% was not significant ($t$(99) = 0.89, $p$ = 0.19). In the unbalanced analysis, the best performing pretrained approach yielded exploratively significant improvements against the unbalanced standard approach ($t$(99) = 2.04, $p$ = 0.02, see Table A4 for details). Contrary to expectations, the approach that simulated 6 features with 20% weight was descriptively inferior to the other approaches. See Figure 1 for a graphical representation of balanced accuracies on both balanced and unbalanced data. The exploratory inclusion of the MADRS improved predictive performance for the pretrained approaches and led to the descriptively best observed performance (see Table A4).

## Interim Discussion

Since both null hypotheses were retained, the utility of the proposed pretraining cannot be assumed, even though we observed descriptive improvement for most pretraining approaches. As prediction accuracies in clinical psychology are typically low (.63 to .82; Sajjadian et al., 2021; Vieira et al., 2022), even minor improvements may be worthwhile. Therefore, even such non-significant improvements could provide a helpful tool to mitigate the lack of shared data in clinical psychology.

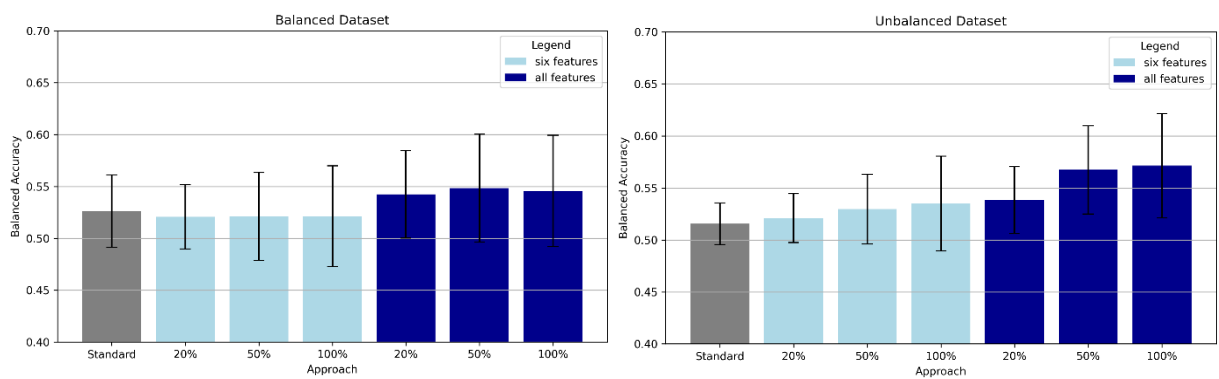

**Figure 1** *Balanced accuracies for response prediction on balanced and unbalanced data – Study 1*

While balancing the data increased the predictive ability of the standard approach, possibly reducing overfitting, which was apparent in a very good train but bad test split performance of the unbalanced standard approach, it reduced the improvement of the pretrained approaches. Both findings might be explained by the change in the specificity values, which were increased by both the pretraining and balancing. Therefore, it seems plausible that one advantage of the simulated data was to provide information about the non-responder class. In addition, while the weight assigned to the simulated data did not greatly impact performance in the balanced analyses, higher weight generally increased predictive accuracy in the unbalanced dataset. This is probably due to the fact that the unbalanced real data provided almost no predictive information, making the simulated data valuable in comparison.

Several limitations still impacted the generalizability of our results. Most importantly, the predictive performance of all classifiers was considerably weaker than reports from the literature (Vieira et al., 2022). While low performance in itself is not a limitation, we suspect that the standard algorithm overfitted on the training set in our analyses. In other publications, overfitting is usually less pronounced which limits the generalizability of our findings, since pretraining might have only mitigated the standard algorithms overfitting. The observed performance should therefore be interpreted carefully. Additionally, while we conducted an extensive literature search to identify studies with descriptive information, the search was not systematic, which might have prevented us from identifying all relevant records.

**Table 1** *Performance metrics for approaches with SMOTE applied – Study 1*

|  | ACC$_{Bal}$ Mean | ACC$_{Bal}$ SD | AUC Mean | AUC SD | Sens Mean | Sens SD | Spec Mean | Spec SD | p H1 |
|---|---|---|---|---|---|---|---|---|---|
| *Standard* | 0.526 | 0.035 | 0.606 | 0.048 | 0.924 | 0.036 | 0.128 | 0.066 | - |
| *six features simulated, 20% weight* | 0.521 | 0.031 | 0.600 | 0.055 | 0.921 | 0.035 | 0.120 | 0.062 | - |
| *six features simulated, 50% weight* | 0.521 | 0.042 | 0.578 | 0.061 | 0.878 | 0.041 | 0.165 | 0.086 | - |
| *six features simulated, 100% weight* | 0.521 | 0.049 | 0.554 | 0.062 | 0.743 | 0.066 | 0.300 | 0.097 | - |
| *all features simulated, 20% weight* | 0.542 | 0.042 | 0.594 | 0.056 | 0.850 | 0.056 | 0.235 | 0.089 | - |
| ***all features simulated, 50% weight*** | **0.549** | **0.052** | **0.588** | **0.057** | **0.680** | **0.132** | **0.417** | **0.145** | **0.19** |
| *all features simulated, 100% weight* | 0.546 | 0.054 | 0.583 | 0.058 | 0.420 | 0.210 | 0.671 | 0.186 | - |

*Note.* The best performing model is printed in bold. ACC$_{Bal}$ = Balanced accuracy, AUC = area under the curve, Sens = Sensitivity, Spec = Specificity, H1 = is the approach superior to the standard algorithm.

### Study 2: improving the prediction of remission due to psychotherapy

To address the limitations of Study 1, we conducted a second analysis on the same dataset with remission instead of response as the label. Here we applied a systematic literature search to identify the studies utilized for simulation.

### Methods

The real dataset was the same as in Study 1. Since our criterion was remission, we conducted a systematic literature search to identify studies that provide descriptive information on remitter/non-remitter OCD patients that were treated with CBT. The variables of interest and criteria for study inclusion were (except for the change from response to remission) the same as in Study 1. The databases Pubmed, Psychinfo and ClinicalTrials.gov (for search terms see Appendix 3) were searched on October 10th, 2024. The studies identified through this search were screened and extracted by NJ. The subsequent data simulation, machine learning and evaluation was comparable to the methodology of Study 1. Since our results regarding the weight for the simulated data were inconclusive, we registered only one hypothesis (https://osf.io/se47c), which states that the best performing pretrained approach will outperform the standard approach in terms of balanced accuracy.

## Results

The systematic literature search yielded 254 references of which 132 were duplicates. Only seven of the 188 remaining studies provided suitable information for the simulation process. Due to that scarcity of descriptive information for OCD remitters/non-remitters we were only able to simulate five variables from the literature information. This resulted in two simulated datasets, one only simulating the five features and one simulating all features with the aforementioned five on the basis of the literature. While the balanced accuracy of the standard algorithm increased due to the altered criterion, the performance of the pretrained approaches diminished dramatically. See Table 2 for an overview of the performance metrics on the balanced data and Figure 2 for a comparison of accuracy in balanced and unbalanced analyses. Also consider Table A5 for all performance metrics of the approaches on the unbalanced dataset.

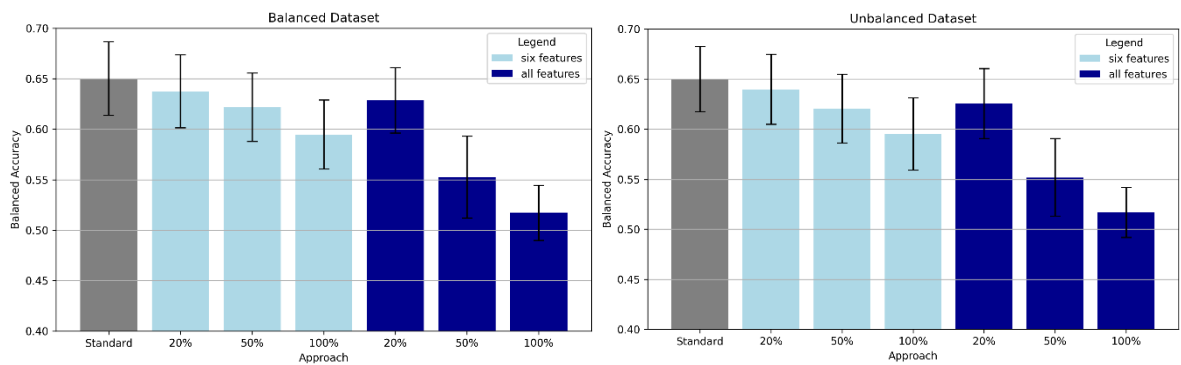

**Figure 2** *Balanced accuracies for remission prediction on balanced and unbalanced data – Study 2*

## General Discussion

**Findings of Study 2**

Taken together, the results do not support the utility of the proposed pretraining method. While we observed a descriptive improvement in predictive performance in study 1, we found a pronounced decrease in Study 2. These results might be impacted by several factors. First, the simulation process in both studies was very simple, relying on the assumption of multivariate normality and did not model nonlinear interactions between the simulated variables. Additionally, only few studies reported the descriptive statistics necessary even for this simple simulation method. This restricted the utilizable information considerably. In Study 2 where, in comparison, even less information was available, the performance of the pretrained approaches was very low, hinting at the impact of information quantity on simulation quality.

**Integration of Findings**

The decreased performance in Study 2 contradicts the descriptive improvement in Study 1 and leaves an unclear picture. For both studies, it is possible that specific dataset characteristics lead to the observed performance. No definitive conclusion can therefore be derived for the approach. When analysing the literature information, underlying our simulations, one might infer the importance of sufficient quality and quantity of this information. The reduced amount of information, incorporated in Study 2, both in terms of number of variables simulated as well as number of studies informing the simulation, may be a major reason for the observed decrease in performance. With more incorporated information and variables, performance was considerably improved in the unbalanced analyses in Study 1, while balancing the dataset had a negligible effect. Therefore, the method might provide some benefit in cases where algorithms display very weak performance when trained on real data, which cannot be improved by application of oversampling.

**Table 2** *Performance metrics for approaches with balanced data – Study 2*

|  | $ACC_{Bal}$ Mean | $ACC_{Bal}$ SD | AUC Mean | AUC SD | Sens Mean | Sens SD | Spec Mean | Spec SD | p H1 |
|---|---|---|---|---|---|---|---|---|---|
| ***Standard*** | **0.65** | **0.036** | **0.709** | **0.035** | **0.63** | **0.062** | **0.67** | **0.059** | - |
| *five features simulated, 20% weight* | 0.637 | 0.036 | 0.695 | 0.036 | 0.581 | 0.058 | 0.694 | 0.055 | 1 |
| *five features simulated, 50% weight* | 0.622 | 0.034 | 0.669 | 0.039 | 0.518 | 0.068 | 0.725 | 0.056 | - |
| *five features simulated, 100% weight* | 0.595 | 0.034 | 0.637 | 0.042 | 0.448 | 0.084 | 0.741 | 0.06 | - |
| *all features simulated, 20% weight* | 0.628 | 0.032 | 0.706 | 0.035 | 0.787 | 0.074 | 0.47 | 0.097 | - |
| *all features simulated, 50% weight* | 0.553 | 0.04 | 0.704 | 0.034 | 0.936 | 0.062 | 0.17 | 0.129 | - |
| *all features simulated, 100% weight* | 0.517 | 0.027 | 0.699 | 0.034 | 0.984 | 0.032 | 0.05 | 0.08 | - |

*Note.* The best performing model is printed in bold. $ACC_{Bal}$ = Balanced accuracy, AUC = area under the curve, Sens = Sensitivity, Spec = Specificity.

**Future research**

While we did not find significant improvement due to pretraining, the problem of data scarcity in clinical psychology remains an important issue. With this paper we aim to present our findings in order to inform potential adaptations and hypotheses. For potential adaptations, some important points ought to be considered while planning the investigation. First, the pretraining should only be applied if a considerable amount of descriptive information regarding the target groups is reported in the literature. This might be the case for commonly investigated disorders (e.g., unipolar affective disorders) but less so for others. Second, our exploratory analyses, which added the MADRS as a simulated variable, yielded improved performance. This suggests that it might be beneficial to simulate (with consideration of time and effort) rather more than less potentially informative variables, which should be carefully selected based on the current evidence (e.g., Kathmann et al., 2022). It seems that strict inclusion criteria for studies should be weighted against information utilization. Also, the approach seems to perform best as a substitute for balancing to increase performance and a sequential approach (i.e., observing performance and effect of oversampling and in case of bad oversampling performance search for information and simulation of data to explore pretraining effects) might be considered. Finally, the assumption of multivariate normality could be substituted by a empirically identified distribution, for example via kernel density estimation (Rosenblatt, 1956).

As an appeal to researchers performing RCT´s and other interventional studies, we encourage the publication of descriptive data for participant subgroups. This could not only broaden the utilizable information for our proposed approach but also contribute to a more nuanced understanding of patient populations.

Overall, the presented pretraining method for random forests yielded inconclusive results. Since other options to utilize data in psychology to a larger extent seem to be rarely used, it might still be interesting to explore its properties on other samples and refine its methodology to circumvent the relative lack of large open datasets that impedes the advance of personalized treatments.

**Acknowledgements:**

The authors wish to thank Larissa Hofert for helpful comments.

**Conflict of Interest statement.**
K.H. is a scientific advisor to the Aury Care GmbH, which develops Aury, an AI-based chatbot providing mental health support. He holds virtual stock options in Aury Care GmbH. Previously, these options were held in the Mental Tech GmbH, from which the Aury Care GmbH took over product development. All other authors report no potential conflicts of interest.

**Author contributions**

NJ, MV, KH conceptualization; NJ, data analysis; NJ, writing - original draft, NJ, MV, NK, KH, writing – review and editing; MV, NK, KH, supervision

**Ethical approval**

The data was collected as part of the routine assessment of the specialty OCD outpatient clinic of the Humboldt-Universität zu Berlin, Germany and met the ethical standards of the revised Declaration of Helsinki.

**Consent for publication**

All participants provided written informed consent that the data being collected during their therapies will be used for research and might be published.

**Availability of data and Material**

The dataset contains sensitive information and can not be easily shared. The analysis scripts are available on GitHub.

# Appendix

**Tables**

**Table A1**

*Studies utilized for mean and sd simulation in Study 1*

|  |  | Variable | | | | | |
|---|---|---|---|---|---|---|---|
| Study | n | YBOCS | BDI | GAF | OCI-R | Onset | Age |
| Mataix-Cols et al. (2022)** | 502 |  |  | • | • |  |  |
| Vandborg et al. (2016) | 39 | • |  | • |  | • | • |
| Olatunji et al. (2014)* | 12 | • | • |  |  |  | • |
| Thiel et al. (2014) | 67 | • | • |  |  | • |  |
| Rufer et al. (2006) | 43 |  | • |  |  |  |  |
| Moritz et al. (2004) | 53 | • | • |  |  |  | • |

*Note.* YBOCS: Yale-Brown Obsessive-Compulsive Scale, BDI II: Beck depression Inventory II, GAF: Global Assessment of Functioning, OCI-R: Obsessive-Compulsive Inventory - Revised, Onset: Age at onset of OCD

* All patients were reported so desired values were obtained directly from the data

** Study did not report numerical values but kindly shared the information upon request

**Table A2**

*Studies utilized for covariance simulation*

| Variable | YBOCS | BDI II | GAF | OCI-R | Age | Onset |
|---|---|---|---|---|---|---|
| YBOCS | | | | | | |
| BDI II | 2, 3, 5, 7, 8, 10, 11, 12, 19 | | | | | |
| GAF | 8, 9, 13, 17 | 8 | | | | |
| OCI-R | 1, 3, 4, 6, 15, 16, 19 | 1, 3, 6, 14, 16, 19 | - | | | |
| Age | 4, 9, 10, 17, 18, 19 | 10, 11, 19 | 17 | 4, 19 | | |
| Onset | 5, 17, 18 | 5 | 17 | - | 17, 18 | |

*Note.* 1: Abramowitz & Deacon (2006), (n=167); 2: Cromer et al. (2007), (n=265); 3: Sulkowski et al. (2008), (n=112); 4: Storch et al. (2010), (n=130); 5: Da Rocha et al. (2011), (n=122); 6: Belloch et al. (2013), (n=90/88); 7: Havnen et al. (2013), (n=6); 8: Williams et al. (2013), (n=75); 9: Hiranyatheb et al. (2014), (n=41); 10: Olatunji et al. (2014), (n=12); 11: Rees et al. (2014), (n=322); 12: Semiz et al. (2014), (n=120); 13: Saipanish et al. (2015), (n=49); 14: Tellawi et al. (2016), (n=63); 15: Hedman et al. (2017), n(=95); 16: Hon et al. (2019), (n=130/113); 17: Chabardes et al. (2020), (n=19); 18: Naesström et al. (2021), (n=11); 19: Paul et al. (2022), (n=46)

**Table A3** *Performance metrics for response classifiers across approaches with balanced data – Study 1*

|  | ACC$_{Bal}$ Mean | ACC$_{Bal}$ SD | AUC Mean | AUC SD | Sens Mean | Sens SD | Spec Mean | Spec SD | p H1 |
|---|---|---|---|---|---|---|---|---|---|
| *Standard* | 0.526 | 0.035 | 0.606 | 0.048 | 0.924 | 0.036 | 0.128 | 0.066 | - |
| *six features simulated, 20% weight* | 0.521 | 0.031 | 0.600 | 0.055 | 0.921 | 0.035 | 0.120 | 0.062 | - |
| *six features simulated, 50% weight* | 0.521 | 0.042 | 0.578 | 0.061 | 0.878 | 0.041 | 0.165 | 0.086 | - |
| *six features simulated, 100% weight* | 0.521 | 0.049 | 0.554 | 0.062 | 0.743 | 0.066 | 0.300 | 0.097 | - |
| *all features simulated, 20% weight* | 0.542 | 0.042 | 0.594 | 0.056 | 0.850 | 0.056 | 0.235 | 0.089 | - |
| *all features simulated, 50% weight* | 0.548 | 0.052 | 0.588 | 0.057 | 0.680 | 0.132 | 0.417 | 0.145 | 0.19 |
| *all features simulated, 100% weight* | 0.546 | 0.054 | 0.583 | 0.058 | 0.420 | 0.210 | 0.671 | 0.186 | - |
| *seven features simulated, 20% weight* | 0.533 | 0.041 | 0.606 | 0.055 | 0.869 | 0.047 | 0.199 | 0.080 | - |
| *seven features simulated, 50% weight* | 0.556 | 0.046 | 0.595 | 0.053 | 0.687 | 0.074 | 0.424 | 0.094 | **-** |
| *seven features simulated, 100% weight* | 0.555 | 0.051 | 0.577 | 0.056 | 0.403 | 0.079 | 0.707 | 0.102 | - |
| *all features simulated, 20% weight, MADRS included* | 0.537 | 0.041 | 0.596 | 0.052 | 0.850 | 0.056 | 0.226 | 0.086 | - |
| ***all features simulated, 50% weight, MADRS included*** | **0.554** | **0.049** | **0.590** | **0.056** | **0.677** | **0.138** | **0.432** | **0.149** | **-** |
| *all features simulated, 100% weight, MADRS included* | 0.548 | 0.050 | 0.585 | 0.057 | 0.406 | 0.206 | 0.690 | 0.102 | - |

*Note.* The best performing model is printed in bold. ACC$_{Bal}$ = Balanced accuracy, AUC = area under the curve, Sens = Sensitivity, Spec = Specificity.

**Table A4** *Performance metrics for response classifiers across approaches with unbalanced data – Study 1*

| | ACC$_{Bal}$ Mean | ACC$_{Bal}$ SD | AUC Mean | AUC SD | Sens Mean | Sens SD | Spec Mean | Spec SD | p H1 |
|---|---|---|---|---|---|---|---|---|---|
| *Standard* | 0.516 | 0.020 | 0.651 | 0.054 | 0.991 | 0.011 | 0.038 | 0.040 | - |
| *six features simulated, 20% weight* | 0.521 | 0.024 | 0.640 | 0.056 | 0.987 | 0.016 | 0.055 | 0.046 | - |
| *six features simulated, 50% weight* | 0.530 | 0.034 | 0.606 | 0.060 | 0.958 | 0.028 | 0.101 | 0.072 | - |
| *six features simulated, 100% weight* | 0.535 | 0.046 | 0.573 | 0.063 | 0.840 | 0.057 | 0.230 | 0.093 | - |
| *all features simulated, 20% weight* | 0.533 | 0.032 | 0.655 | 0.052 | 0.974 | 0.022 | 0.093 | 0.060 | - |
| *all features simulated, 50% weight* | 0.564 | 0.042 | 0.645 | 0.054 | 0.877 | 0.075 | 0.251 | 0.116 | - |
| *all features simulated, 100% weight* | 0.571 | 0.050 | 0.633 | 0.058 | 0.561 | 0.227 | 0.578 | 0.093 | 0.02 |
| *seven features simulated, 20% weight* | 0.530 | 0.027 | 0.656 | 0.053 | 0.979 | 0.019 | 0.080 | 0.051 | - |
| *seven features simulated, 50% weight* | 0.552 | 0.044 | 0.634 | 0.056 | 0.879 | 0.051 | 0.226 | 0.079 | - |
| *seven features simulated, 100% weight* | 0.568 | 0.052 | 0.602 | 0.061 | 0.542 | 0.0912 | 0.595 | 0.106 | - |
| *all features simulated, 20% weight, MADRS included* | 0.534 | 0.031 | 0.652 | 0.054 | 0.975 | 0.021 | 0.094 | 0.060 | - |
| *all features simulated, 50% weight, MADRS included* | 0.567 | 0.041 | 0.645 | 0.056 | 0.879 | 0.082 | 0.255 | 0.125 | - |
| **all features simulated, 100% weight, MADRS included** | **0.573** | **0.054** | **0.631** | **0.058** | **0.564** | **0.230** | **0.583** | **0.106** | **-** |

*Note.* The best performing model is printed in bold. ACC$_{Bal}$ = Balanced accuracy, AUC = area under the curve, Sens = Sensitivity, Spec = Specificity.

**Table A5** *Performance metrics for approaches with unbalanced data – Study 2*

|  | ACC<sub>Bal</sub> Mean | ACC<sub>Bal</sub> SD | AUC Mean | AUC SD | Sens Mean | Sens SD | Spec Mean | Spec SD | p |
|---|---|---|---|---|---|---|---|---|---|
| ***Standard*** | **0.65** | **0.033** | **0.708** | **0.036** | **0.621** | **0.066** | **0.679** | **0.061** | - |
| *five features simulated, 20% weight* | 0.64 | 0.035 | 0.696 | 0.037 | 0.574 | 0.067 | 0.705 | 0.057 | 1 |
| *five features simulated, 50% weight* | 0.621 | 0.034 | 0.672 | 0.04 | 0.509 | 0.072 | 0.732 | 0.056 | - |
| *five features simulated, 100% weight* | 0.595 | 0.036 | 0.64 | 0.043 | 0.445 | 0.086 | 0.745 | 0.063 | - |
| *all features simulated, 20% weight* | 0.626 | 0.035 | 0.708 | 0.034 | 0.783 | 0.08 | 0.468 | 0.104 | - |
| *all features simulated, 50% weight* | 0.552 | 0.039 | 0.706 | 0.035 | 0.934 | 0.072 | 0.17 | 0.136 | - |
| *all features simulated, 100% weight* | 0.517 | 0.025 | 0.703 | 0.035 | 0.984 | 0.039 | 0.05 | 0.083 | - |

*Note.* The best performing model is printed in bold. ACC<sub>Bal</sub> = Balanced accuracy, AUC = area under the curve, Sens = Sensitivity, Spec = Specificity.

**Table A6** *Descriptive Statistics of the Simulated features for responders/non-responders – Study 1*

| Variable | Mean_Responder | SD_Responder | Mean_Non-responder | SD_Non-responder |
|---|---|---|---|---|
| YBOCS | 25.38 | 5.60 | 24.96 | 5.82 |
| BDI II | 16.80 | 9.33 | 22.53 | 11.20 |
| GAF | 59.11 | 7.08 | 57.85 | 8.28 |
| OCI-R | 26.07 | 12.12 | 26.47 | 12.93 |
| Onset | 18.92 | 8.00 | 16.57 | 8.76 |
| Age | 30.43 | 8.98 | 33.61 | 10.76 |

*Note.* YBOCS: Yale-Brown Obsessive-Compulsive Scale, BDI II: Beck depression Inventory II, GAF: Global Assessment of Functioning, OCI-R: Obsessive-Compulsive Inventory - Revised, Onset: Age at onset of OCD.

**Table A7** *Descriptive Statistics from the real data for the Variables of interest for responders/non-responders – Study 1*

| Variable | Mean$_{Responder}$ | SD$_{Responder}$ | Mean$_{Non-responder}$ | SD$_{Non-responder}$ |
|---|---|---|---|---|
| YBOCS | 23.70 | 5.12 | 20.84 | 5.97 |
| BDI II | 18.63 | 10.96 | 18.05 | 10.19 |
| GAF | 55.46 | 10.22 | 56.12 | 10.05 |
| OCI-R | 27.36 | 11.67 | 27.40 | 13.20 |
| Onset | 18.14 | 9.74 | 18.28 | 8.83 |
| Age | 32.85 | 10.62 | 32.94 | 9.11 |

*Note.* YBOCS: Yale-Brown Obsessive-Compulsive Scale, BDI II: Beck depression Inventory II, GAF: Global Assessment of Functioning, OCI-R: Obsessive-Compulsive Inventory - Revised, Onset: Age at onset of OCD.

**Appendix 1: Extraction of information and simulation**

The available literature was searched for studies that reported suitable descriptive statistics for the YBOCS (Goodman, 1989), Beck Depression Inventory-II (BDI-II; Beck et al., 1996a), global assessment of functioning (GAF; APA, 1994), Obsessive Compulsive Inventory – Revised (OCI-R; Foa et al., 2002), Montgomery-Åsberg Depression Rating Scale (MADRS; Montgomery & Åsberg, 1979), Age at the beginning of treatment and Age of onset of disorder. These features were selected due to the anticipation that they would be predictive for the task at hand, based on the analyses by Hilbert et. al. (2021). The desired statistics included the means and standard deviations, as well as the correlation coefficients between the features. These were necessary to obtain a mean vector and a variance-covariance matrix for the simulation process afterwards.

Requirements for studies to be used for mean and standard deviation extraction were that they (a) reported adult OCD patients that did not exclusively exhibit severe OCD; (b) reported mean and standard deviation for responders and non-responders; and (c) their applied treatment was CBT. These requirements were employed to obtain coefficients for simulation from patients that closely matched the participants in the real dataset. Only six out of several hundred screened studies met our inclusion criteria (see Table A1 for details). For the MADRS, no study fulfilled all requirements, therefore it was not simulated in the main analyses (six features dataset). It was however exploratively simulated and analyzed in separate (seven features) dataset which encompassed the MADRS in addition to the other features of interest. This was done to illustratively examine whether it is beneficial to include more variables or adhere to strict criteria. The mean and standard deviation coefficients for this illustrative MADRS inclusion were obtained from Kim et al. (2021). This study applied only pharmacological treatment and thereby violated requirement c for studies reporting means and standard deviations.

Regarding covariances between the variables of interest our search identified no studies that differentiated covariances between responders and non-responders. Therefore, the requirements were less strict for studies that reported correlations to facilitate the incorporation of information from the literature. Studies from which correlation coefficients were obtained only had to report values obtained from adult OCD patients. The studies utilized for the simulation of the correlations are displayed in Table A2. For the explorative simulation of the MADRS, some studies used for correlation extraction met the requirement (Hedman et al., 2017; Naesström et al., 2021; Paul et al., 2022) while others reported patients with a primary diagnosis of depression (Benazzi, 1999; Wikberg et al., 2015). Covariances that could not be obtained from the literature were generally substituted by the descriptive covariance in the train split of the data, which is only used for fitting the models (for an overview of covariance studies see Table A2).

Some studies reported the BDI (Beck et al., 1961) instead of the BDI-II or YBOCS II (Storch et al., 2010) instead of YBOCS. However, since the measures correlate closely with .93 (Beck et al., 1996b; Dozois et al., 1998) for the BDI-I/II and .97 for the YBOCS I/II (Storch et al., 2010), their results were treated as if they had reported the BDI-II/YBOCS. This approach was chosen, since the gathered statistics overall stemmed from diverse samples and different translated versions, rendering this minor distinction relatively unimportant in comparison. In total, the cumulated values from the literature stemmed from multiple patients with different backgrounds and nationalities, therefore providing a heterogeneous sample. The descriptive statistics of the simulated and real features of interest are displayed in Table A6/A7.

The descriptive statistics obtained from the literature were used to simulate the data for pretraining. To achieve this, in a first step, the coefficients were weighted according to their respective sample size. If, for instance, the BDI-II was found to have a responder mean of 18 in one study (n=20) and of 24 in another (n=40), this would result in a weighted mean of 22. Here the sample size of the overall sample, not the size of the corresponding group (responder/non-responder) was used in order not to distort the observed associations. Relying on group sizes would enable a situation where a study reported overall high BDI-II values but a disproportionately higher number of responding than non-responding patients. This would artificially elevate the weighted mean of responders, while the non-responder mean would remain less affected. Consequently, the overall weighted mean of responders could be higher than that of the non-responders contradicting the lower mean observed in each study individually. In addition to weighting, the standard deviations were squared, to obtain estimates of the variance which were used to transform the obtained correlation coefficients into covariances.

The obtained weighted means, variances, and covariances were then used to define two multivariate normal distributions (MVND) $N(\mu,\Sigma)$ where $\mu$ is a mean vector and $\Sigma$ is a variance-covariance matrix. This was done using the multivariate_normal function from the NumPy package (Harris et al., 2020). The first MVND was defined by the obtained parameters for responding patients, while the second MVND used the parameters for non-responders. The correlation coefficients used for both distributions were identical although they still resulted in slightly different covariances as the variances used to calculate them were responder/non-responder specific. From both MVNDs 500 cases were drawn, which received their label (1 = responder, 0 = non-responder) based on the MVND they were sampled from. Lastly, the two samples were combined and randomly shuffled.

Two different simulation approaches were utilized. The first one simulated the features of interest obtained from the literature based on their means, variances, and covariances (six/seven features). For the second the same features of interest were simulated, additionally, all other features in the real data were simulated as well, using their means, variances, and covariances in the train split (all features). This was done to explore the influence of the feature space restriction that arises from the six/seven features approach. Both methods yielded a sample with 1000 cases, of which 500 were simulated as responders and 500 as non-responders, which were used for pretraining. The 50/50 ratio was chosen to enable the model to infer a predictive model that performs equally well for both categories.

**Appendix 2: ML pipeline**

In order to be able to separately examine the performance of the different approaches, the analyses were performed independently from each other. To discover possible improvements in performance, a comparison to a random forest that was trained only on the real data was performed.

Since 72% of patients were categorized as responders, the synthetic minority oversampling technique (SMOTE; Chawla et al., 2002) was applied to the training datasets. SMOTE uses a ML algorithm to create synthetic samples that resemble the structure prevalent in the minority class. These synthetic samples are added to the dataset until both outcome classes contain the same number of cases. Specifically, SMOTENC from the imbalanced-learn library (Lemaître et al., 2017) was utilized. An explorative analysis in which SMOTE was not applied was performed so that a possible improvement for unbalanced datasets via pretraining could be investigated. Since 4 datasets were simulated, each of which was combined once with the unbalanced and once with the balanced real data for fine-tuning, 8 separate analyses were carried out.

All random forests received certain hyperparameters which were tuned via a fivefold grid search to approximate realistic scenarios. This hyperparameter-tuning was conducted separately for all datasets since each dataset could have potentially required different hyperparameters for optimal classifier performance. In the following, the tuned hyperparameters are shortly described. First, the maximum number of features used for the construction of a single decision tree ($\sqrt{n_{\text{features}}}$, $\log_2 n_{features}$, 4, 5) was tuned. The 4 features and 5 features options were allowed to enable the usage of a substantial portion of features since some of the simulated datasets only contained 6 or 7 features. Also, the minimum number of samples in each leaf node (1, 3, 5, 10) and the maximum number of samples used for construction (66%, 80%, 100%) were tuned. The total number of decision trees trained was equal for each approach and set to 200.

After preprocessing, the random forests were trained on their respective datasets (depending on the weight parameter more or less trees were trained on the simulated data). For the pretrained approaches, the probabilistic predictions of all trees (trained on simulated and trained on real data) for the test set were combined into one ensemble and averaged. If the averaged probability was > 0.5 the final decision of the random forest was 1, otherwise 0. This voting method is consistent with the procedure implemented in the machine learning python library scikit learn (Pedregosa et al., 2011). Since the scikit learn implementation of the random forest was utilized for the standard algorithm, this ensured comparability between both approaches.

The data splitting, subsequent data simulation, preprocessing, hyperparameter tuning, balancing, training of the pretrained and standard forests, and the evaluation on the respective test split were repeated for 100 iterations (Monte Carlo Cross Validation, MCCV). The MCCV was performed to reduce the probability of performance improvement by chance. For each iteration, several performance metrics were saved. The approaches integrating pretrained and fine-tuned trees (pretrained random forests) were primarily compared to the standard random forests in terms of balanced accuracy, which is defined as the mean of sensitivity and specificity. This metric was employed because it is a commonly used tool in ML contexts and assigns every class of labels the same weight, thereby avoiding the higher weight for the majority class given by the regular accuracy (Grandini et al., 2020). Additionally, the sensitivity, specificity, and the area under the receiver operator characteristic curve (AUROC/AUC) were calculated. Still, the balanced accuracy was used as the primary metric to compare models here because the performance of the model with the implemented threshold was the most interesting. This is the case because, in a precision medicine scenario, it is likely that one specified cut-off, which showed the best performance in the training of the algorithm, will be used for prediction.

**Appendix 3: Search terms**

The search terms in our systematic search for descriptive information about remitters and non-remitters were („Obsessive compulsive disorder" OR „OCD") AND („remitters" OR „responders" OR „remitted patients" OR „responding patients") AND („YBOCS" OR „Yale-Brown Obsessive-Compulsive Scale" OR „BDI II" OR „BDI-II" OR "BDI" OR „Beck Depression Inventory" OR „GAF" OR „Global Assessment of Functioning" OR „OCI-R" OR „Obsessive-Compulsive Inventory – Revised" OR „Age at onset" OR "Age of onset" OR „Age" OR „MADRS" OR „Montgomery-Åsberg Depression Rating Scale") AND ("Cognitive behavioral therapy" OR „CBT" OR „exposure" OR „behavioral therapy").